\documentclass[sigconf]{acmart}

\AtBeginDocument{%
  \providecommand\BibTeX{{%
    \normalfont B\kern-0.5em{\scshape i\kern-0.25em b}\kern-0.8em\TeX}}}

\setcopyright{acmcopyright}
\copyrightyear{2023}
\acmYear{2023}
\acmDOI{XXXXXXX.XXXXXXX}

\acmConference[XXXXXX 2023]{Conference XXXXXX 2023}{ }{ }
\acmPrice{}
\acmISBN{}

\acmSubmissionID{4676}

\usepackage{microtype}
\usepackage{mathtools}

\usepackage{subfigure}  
\usepackage{algorithmicx}  
\usepackage{algpseudocode}  
\usepackage{amsmath}
\usepackage{algorithm}    

\usepackage{amsthm}

\makeatletter
\DeclareRobustCommand\onedot{\futurelet\@let@token\@onedot}
\def\@onedot{\ifx\@let@token.\else.\null\fi\xspace}
\def\eg{\emph{e.g}\onedot} 
\def\ie{\emph{i.e}\onedot}

\def\etal{\emph{et al}\onedot}

\usepackage{xspace}
\newcommand{\name}[0]{NCN\xspace}

\usepackage{xcolor}

\begin{document}




\title{Neighborhood Convolutional Network: A New Paradigm of Graph Neural Networks for Node Classification}




\author{Jinsong Chen$^{1,2,\ast}$, Boyu Li$^{1,\ast}$, Kun He$^{1,\dag}$}

\thanks{$\ast$ Both authors contributed equally to this research.}
\thanks{$\dag$ Corresponding author.}
\affiliation{%
  \institution{$^{1}$ School of Computer Science and Technology, Huazhong University of Science and Technology   
  \city{Wuhan}
  \country{China}}
  \institution{$^{2}$ Institute of Artificial Intelligence, Huazhong University of Science and Technology
  \city{Wuhan}
  \country{China}}
}
\email{{chenjinsong,afterslby,brooklet60}@hust.edu.cn}

\def\authors{Jinsong Chen, Boyu Li, Kun He}

\renewcommand{\shortauthors}{Chen, et al.}

\begin{abstract}
The decoupled Graph Convolutional Network (GCN), a recent development of GCN that decouples the neighborhood aggregation and feature transformation in each convolutional layer, has shown promising performance for graph representation learning. 
Existing decoupled GCNs first utilize a simple neural network (\eg, MLP) to learn the hidden features of the nodes, 
then propagate the learned features on the graph with fixed steps to aggregate the information of multi-hop neighborhoods.
Despite effectiveness, the aggregation operation, which requires the whole adjacency matrix as the input, is involved in the model training, causing high training cost that hinders its potential on larger graphs.
On the other hand, due to the independence of node attributes as the input, the neural networks used in decoupled GCNs are very simple, and advanced techniques cannot be applied to the modeling. 
To this end, we further liberate the aggregation operation from the decoupled GCN and propose a new paradigm of GCN, termed Neighborhood Convolutional Network (\name), that utilizes the neighborhood aggregation result as the input,
followed by a special convolutional neural network tailored for extracting expressive node representations from the aggregation input. 
In this way, the model could inherit the merit of decoupled GCN for aggregating neighborhood information, at the same time, develop much more powerful feature learning modules. 
A training strategy called mask training is incorporated to further boost the model performance.
Extensive results demonstrate the effectiveness of our model for the node classification task on diverse homophilic graphs and heterophilic graphs.
\end{abstract}



\maketitle

\section{Introduction}
Graphs are ubiquitous in real-world scenarios, such as social networks on social media platforms and molecular structures in biology analysis.
Due to the wide applications of graphs, many data mining tasks on graphs have been proposed and investigated well in the field of graph representation learning. 
As one of the most fundamental and important tasks, node classification, whose goal is to predict node labels, has received considerable attention in recent years.
Among various techniques for node classification, Graph Convolutional Network (GCN)~\cite{gcn}, as a typical model of Graph Neural Network (GNN), has drawn great popularity owing to its promising performance and high flexibility. 

The vanilla GCN~\cite{mpnn}, consisting of a few GCN layers, aggregates the topology features and attribute features of immediate neighbors to learn the representations of nodes. 
There are two main modules, neighborhood aggregation and feature transformation~\cite{pt}, in each GCN layer. 
First, the neighborhood aggregation module collects the features of the immediate neighbors of each node in the graph through the adjacency matrix. 
Then, the feature transformation module utilizes a linear layer to learn the hidden representation of each node based on the aggregated information. 
Neighborhood aggregation and feature transformation are always coupled in each layer of the GCN frameworks~\cite{gcnii,gdc,jknet,ugcn}.
Nevertheless, recent studies illustrate that such coupled design would cause high training cost~\cite{appnp} and lead to over-smoothing issue~\cite{sgcn} if stacking many GCN layers, limiting the model to capturing deep graph structural information. 

Thereafter, decoupled GCN~\cite{pt} emerges to tackle the above issues. 
Its key idea is to separate the neighborhood aggregation module (a.k.a. the propagation module) and feature transformation module into independent modules.
Decoupled GCN first leverages a neural network (\ie, MLP) as the feature transformation module to learn the hidden representations of nodes.
Then, decoupled GCN propagates the hidden representations of nodes over the graph to extract the final representations of nodes that involve the semantic information of multi-hop neighborhoods.
By proposing various propagation methods~\cite{appnp,sgcn} and aggregation strategies~\cite{deepergcn,gprgnn}, 
decoupled GCNs have shown promising performances for the node classification task, attracting increasing attention in recent years.

Despite effectiveness, we observe there are two main weaknesses in existing decoupled GCNs.
First, the costs of model training, including time and space, are expensive on large graphs, limiting their scalability.
It is because the propagation operation is involved in the training stage, so that the whole adjacency matrix has to be utilized in every training epoch.
Second, the semantic information of multi-hop neighborhoods has not been fully explored yet. 
Decoupled GCNs utilize the normalized adjacency matrix to learn the hidden representations of nodes, which is inefficient in discovering the semantic information of multi-hop neighborhoods.

To this end, we propose a novel and flexible framework called Neighborhood Convolutional Network (\name), which consists of three main stages:
data pre-processing, feature extracting and feature fusion.
For data pre-processing, \name propagates the raw attribute features of nodes and transforms the features of multi-hop neighborhoods into a grid-like feature matrix for each node. 
The grid-like feature matrix brings non-Euclidean space of independent nodes to order, and more advanced deep learning modules can be applied on the grid-like feature matrix to extract more semantic information for nodes.
Then, in the feature extracting stage, \name develops a Convolutional Neural Network (CNN) based module to capture the high-level semantic features from the grid-like feature matrix. 
In this way, \name could extract more semantic information from multi-hop neighborhoods than decoupled GCNs.
Finally, in the feature fusion stage, a learnable weight layer is utilized to learn the aggregation weights of the original attribute features of nodes and the extracted features learned from multi-hop neighborhoods, respectively. 
Thus, \name can adaptively adjust the aggregation weights according to the characteristics of different graphs. 
Moreover, \name develops a mask-based training strategy to further improve the performance of model training.
We conduct extensive experiments on various benchmark datasets, including homophilic graphs and heterophilic graphs, to validate the effectiveness of \name for the node classification task.

The main contributions of this work are summarized as follows:

\begin{itemize}
\item We propose \name, a new and effective paradigm of GCN for the node classification task, that can extract the node representations involving rich semantic information of multi-hop neighborhoods.

\item We utilize CNN based modules to effectively extract the features from multi-hop neighborhoods and apply a learnable weight layer to adaptively learn the aggregation weights of the original attribute features and the extracted features. Besides, a mask-based training strategy is proposed to further boost training performance.

\item We analyze the time complexity and space complexity of \name. Compared to decoupled GCN, \name is more efficient and scalable, especially on large-scale graphs.

\item We conduct extensive experiments for the node classification task on ten benchmark datasets, including homophilic graphs and heterophilic graphs. 
The results demonstrate the superiority of \name over the state-of-the-art baselines.
\end{itemize}

\section{Preliminaries}
\begin{figure*}[t]
    \centering
	\includegraphics[width=17.8cm]{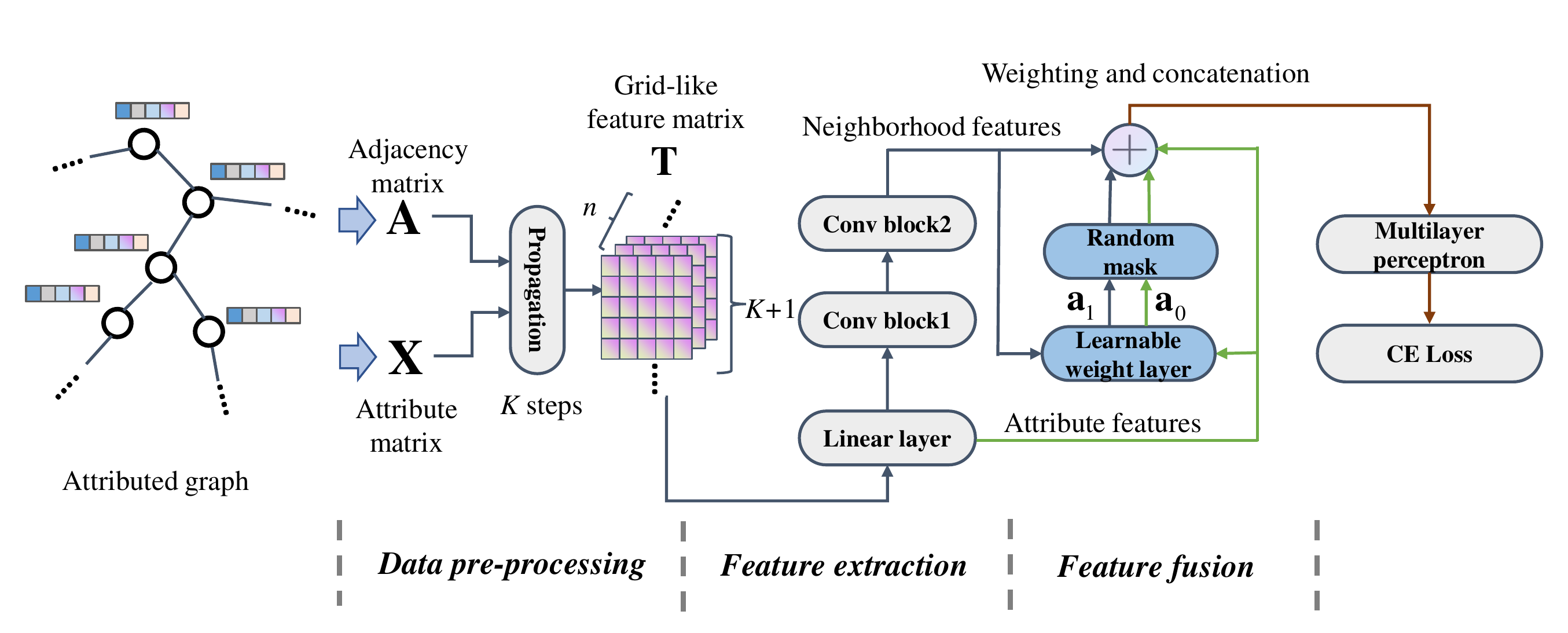}
	\caption{Illustration of \name. There are three main stages. 
	In Stage 1, \name develops a novel aggregator that utilizes the propagation operation to calculate the features of 
    multi-hop neighborhoods and further transform them into the grid-like feature matrix.
	In Stage 2, \name leverages a CNN based module to extract the neighborhood features from the feature matrix. 
	In Stage 3, \name designs a learnable weight layer to combine the features of node attributes and neighborhoods adaptively. A mask-based strategy is applied to boost the model training.}
	\label{fig:fw}
\end{figure*}

\subsection{Problem Formulation}
Given a network modeled as an undirected and attributed graph $G=(\mathcal{V}, \mathcal{E}, \mathbf{X})$, where $\mathcal{V}$ and $\mathcal{E}$ denote the sets of nodes and edges, respectively, and $\mathbf{X}\in \mathbb{R}^{n \times d}$ represents the attribute feature matrix of nodes given that $n$ is the number of nodes and $d$ the dimension of feature vector. 
Let $\mathbf{A}\in \mathbb{R}^{n \times n}$ represent the adjacency matrix, and the normalized adjacency matrix is defined as $\hat{\mathbf{A}} = \mathbf{D}^{-1/2}\mathbf{A} \mathbf{D}^{-1/2}$, where $\mathbf{D}$ is the diagonal degree matrix given as $\mathbf{D}_{ii} = \sum_{j=1}^{n} \mathbf{A}_{ij}$. 
$\mathbf{Y}\in \mathbb{R}^{n \times c}$ denotes the label matrix of nodes, where $c$ denotes the number of labels. 
Given a set of nodes $V_l$ whose labels are known, the node classification task is to predict the labels for nodes $V/V_l$.

\subsection{Homophily and Heterophily}\label{hom}
In this paper, we use the edge homophily ratio $\mathcal{H}(G)$~\citep{h2gnn,bmgcn,gprgnn} to measure the homophily level of a graph $G$, such that
\begin{equation}
	\mathcal{H}(G) = \frac{1}{|V|}\sum_{v_i\in V}\frac{|\{v_j|v_j\in \mathcal{N}_{v_i}, \mathbf{Y}_{v_j}=\mathbf{Y}_{v_i}\}|}{|\mathcal{N}_{v_i}|}, 
	\label{eq:homo}
\end{equation}
where $\mathcal{N}_{v_i}$ denotes the immediate neighbors of node $v_i$. Then, $\mathcal{H}(G) \in [0, 1]$ denotes the degree of homophily, where $\mathcal{H}(G) \to 1$ indicates that $G$ has strong homophily and $\mathcal{H}(G) \to 0$ indicates that $G$ has low homophily (strong heterophily).

\subsection{Graph Convolutional Network}
The key idea of Graph Convolutional Network (GCN) is to apply the first-order approximation of spectral convolution~\citep{sp-gcn} that aggregates the information of immediate neighbors. A GCN layer can be given as
\begin{equation}
	\mathbf{H}^{(l+1)} = \sigma( \hat{\tilde{\mathbf{A}}}\mathbf{H}^{(l)} \mathbf{W}^{(l)}),
	\label{eq:gcn}
\end{equation}
where $\hat{\tilde{\mathbf{A}}}=\tilde{\mathbf{D}}^{-1/2}\tilde{\mathbf{A}}\tilde{\mathbf{D}}^{-1/2}$ represents the normalized adjacency matrix given that $\tilde{\mathbf{A}}$ is the adjacency matrix with self-loops and $\widetilde{\mathbf{D}}$ is the corresponding diagonal degree matrix,
$\mathbf{H}^{(l)} \in \mathbb{R}^{n\times d^{(l)}}$ denotes the representations of nodes in the $l$-th layer,
$\mathbf{W}^{(l)} \in \mathbb{R}^{d^{(l)} \times d^{(l+1)}}$ is a learnable parameter matrix,
and $\sigma(\cdot)$ represents a non-linear activation function.

\subsection{Decoupled Graph Convolutional Network}
According to \autoref{eq:gcn}, a GCN layer contains two coupled operations, namely neighborhood aggregation and feature transformation. 
Such coupled framework could cause the over-smoothing issue~\citep{deepgcn1} and increase the training cost~\citep{sgcn} when stacking many GCN layers.
To address these issues, decoupled GCN~\cite{gprgnn} separates the two operations by removing the activation function and collapsing the parameter matrices between layers.
In this way, the model can be regarded as a combination of an aggregation module (also known as a feature propagation module) and a feature transformation module.
A general framework of decoupled GCN is given as
\begin{equation}
	\mathbf{Z}=\sum_{k=0}^{K} \alpha_{k} \mathbf{H}^{(k)}, \mathbf{H}^{(k)}=\hat{\tilde{\mathbf{A}}} \mathbf{H}^{(k-1)}, \mathbf{H}^{(0)}=\boldsymbol{f}_{\theta}(\mathbf{X}),
	\label{eq:dgcn}
\end{equation}
where $\boldsymbol{f}_{\theta}$ denotes a neural network (\eg, MLP) with a parameter set $\theta$, $\mathbf{H}^{(k)}\in \mathbb{R}^{n\times d^{\prime}}$ and $\alpha_{k}$ represent the representations of nodes and aggregation weight at step $k$, respectively. Here $K$ is the total propagation step, and $\mathbf{Z}\in \mathbb{R}^{n\times d^{\prime}}$ denotes the output of decoupled GCN.

\section{Methodology}

In this section, we propose a novel framework called \textbf{N}eighborhood \textbf{C}onvolutional \textbf{N}etwork (\name), which consists of three main stages, namely data pre-processing, feature extraction, and feature fusion. 
The overall framework of \name is shown in \autoref{fig:fw}, and we detail each stage in the following subsections.

\subsection{Data Pre-processing}
In the data pre-processing stage, we apply a novel aggregator called grid-like neighborhood aggregator (GNA) to preserve the information of multi-hop neighborhoods for each node that can be effectively used for feature extraction. GNA contains two main steps, namely propagating the features of multi-hop neighborhoods and constructing a grid-like feature matrix. 
First, GNA utilizes a propagation operation to aggregate the attribute features of neighborhood nodes, such that
\begin{equation}
	\mathbf{B}^{(k)} = \mathbf{S}^{(k)}\mathbf{X},~~ k \in \{0,1,2,...,K\}, 
	\label{eq:gna}
\end{equation}
where $\mathbf{S}^{(k)} \in \mathbb{R}^{n\times n}$ denotes an aggregation weight matrix at the propagation step $k$, and $\mathbf{S}^{(0)}$ is initialized as the identity matrix $\mathbf{I}_{n}$, such that we consider the nodes themselves as their 0-hop neighborhood node.
$\mathbf{B}^{(k)}\in \mathbb{R}^{n\times d}$ represents the features aggregated from the $k$-hop neighborhood nodes.

In this paper, we apply the approximate personalized PageRank~\citep{appnp,pt} to calculate $\mathbf{S}^{(k)}$,
given as $\mathbf{S}^{(k)}=(1-\gamma)^{k}\hat{\mathbf{A}}^{k} + \gamma\sum_{i=0}^{k-1}(1-\gamma)^{i}\hat{\mathbf{A}}^{i}$, where $\gamma\in{(0,1]}$ is the teleport probability. 
Note that the propagation operation can be implemented by various methods according to the actual demands. For instance, we can set $\mathbf{S}^{(k)}=\hat{\mathbf{A}}^k$ that utilizes the standard random walk to obtain the features of neighborhood nodes.

In the second step, GNA transfers all the features obtained by the first step into a grid-like feature matrix. Specifically, for node $v$, let $\{\mathbf{B}^{(0)}_{v},\mathbf{B}^{(1)}_{v}$ $,\cdots, \mathbf{B}^{(K)}_{v}\}$ represent the features aggregated by performing the propagation operation on the node. GNA constructs a grid-like feature matrix $\mathbf{T}_{v}\in \mathbb{R}^{(K+1)\times d}$ by stacking $\{\mathbf{B}^{(0)}_{v},\mathbf{B}^{(1)}_{v}$ $,\cdots, \mathbf{B}^{(K)}_{v}\}$ in the ascending order according to the propagation step.
In this way, the grid-like feature matrix preserves the feature information of multi-hop neighborhoods from shallow to deep. For instance, the feature vector of the $k$th row in the grid-like feature matrix contains wider extracted information than those of less than $k$th rows. Such design brings non-Euclidean space of independent nodes to order. As a result, more advanced deep learning modules, such as CNN and RNN, can be applied on the grid-like feature matrix to extract more semantic information for nodes.

Note that the whole data pre-processing is non-parametric and can be conducted before the model training process, and therefore it can significantly reduce the training cost. 
Moreover, by assigning an independent feature matrix for each node, the mini-batch training strategy can be efficiently performed on the training model that guarantees the scalability for large-scale graphs.

\subsection{Feature Extraction}
In the feature extraction stage, we capture the complex semantic information from the grid-like neighborhood feature matrix by using a CNN-based module. Specifically, we design a neural network block consisting of two convolutional layers with different kernel sizes. For the first convolutional layer, we set the kernel size as $((\lfloor K/2 \rfloor+1), 1)$ that can capture the semantic association between the features of different hop neighborhoods, where $\lfloor \cdot \rfloor$ denotes the operation of rounding down the value. For the second convolutional layer, the kernel size is set as $(1,1)$ for feature transformation, which is a popular design of CNN in computer vision~\citep{resnet}.

In \name, we first utilize a linear layer to reduce the dimension of the grid-like neighborhood feature matrix since the dimension of attribute features could be very large in some networks.
By stacking two proposed neural network blocks, \name extracts a feature matrix $\mathbf{H}^{N} \in \mathbb{R}^{n \times d^{\prime}}$ that represents the semantic information from multi-hop neighborhoods, given as: 
\begin{equation}
	\mathbf{H}^{N} = \mathrm{Block2}(\sigma(\mathrm{Block1}(\mathbf{T}))), 
	\label{eq:cnn}
\end{equation}
where $\mathbf{T}\in \mathbb{R}^{n\times (K+1)\times d}$ is the grid-like neighborhood feature matrix, $\sigma(\cdot)$ denotes an activation function, and Block1 and Block2 are the two convolutional blocks, respectively.

Note that the design of the neural network block can be flexibly adjusted according to 
different scenarios. In other words, various convolutional layers can be applied to improve the performance of \name. We will study the influence of combining a variety of convolutional layers in our future work.

\subsection{Feature Fusion}
Although the extracted features include rich semantic information of multi-hop neighborhoods, the function of raw attribute features of nodes is seriously weakened.
Recent studies~\citep{h2gnn,gprgnn,glognn} prove that utilizing only the raw attribute features of nodes can also achieve a competitive performance on some graphs. It is because, in some graphs, the raw attribute features of nodes dominate the node labels. Therefore, to accurately predict the labels of nodes, a feasible solution is to learn the representations of nodes by combining the extracted features and the raw attribute features.

To this end, we present an adaptive feature fusion module that can fuse the extracted features and the raw attribute features for different graphs. 
First, we transfer the raw feature matrix $\mathbf{X}$ into a matrix whose shape is the same as the extracted feature matrix, such that:
\begin{equation}
	\mathbf{H}^{R}=\mathbf{XW_{R}}, 
	\label{eq:ra}
\end{equation}
where $\mathbf{W_{R}} \in \mathbb{R}^{d\times d^{\prime}}$ denotes a parameter matrix. Then, we apply a learnable weight layer to adaptively calculate the weights for the feature matrices of $\mathbf{H}^{R}$ and $\mathbf{H}^{N}$, given as:
\begin{equation}
	\mathbf{a} = \mathrm{softmax}(\varphi(\mathbf{H}^{R}||\mathbf{H}^{N})), 
	\label{eq:lwl}
\end{equation}
where $||$ denotes the concatenation operation and $\varphi(\cdot)$ is a linear layer with a parameter matrix $\mathbf{W}_{a}\in \mathbb{R}^{2d^{\prime}\times 2}$ shared by all nodes. The output $\mathbf{a}\in \mathbb{R}^{n\times 2}$ denotes the aggregation weight vector in which $\mathbf{a}_{0}$ and $\mathbf{a}_{1}$ are the aggregation weights of $\mathbf{H}^{R}$ and $\mathbf{H}^{N}$, respectively.
Based on the aggregation weights, the final representation of nodes is calculated as:
\begin{equation}
	\mathbf{H}^{F} = [\mathbf{a}_{0}\odot\mathbf{H}^{R}||\mathbf{a}_{1}\odot\mathbf{H}^{N}],
	\label{eq:fr}
\end{equation}
where $\odot$ denotes the element-wise product.
In practice, the broadcast mechanism of the programming tools (\eg, Pytorch) guarantees the calculation of \autoref{eq:fr} by repeating the columns of $\mathbf{a}_{0}$ and $\mathbf{a}_{1}$ for matching the dimensions of $\mathbf{H}^{R}$ and $\mathbf{H}^{N}$ respectively.

\textbf{Mask Training}
Additionally, inspired by the famous training technique of Dropout~\citep{dropout}, we propose a training strategy called mask training that randomly sets $\mathbf{a}_{0}$ or $\mathbf{a}_{1}$ to zero during the training process. In this way, \name can alleviate the over-fitting problem
during the training. 
Specifically, for each training epoch, we randomly partition the training set into three parts $N_0$, $N_1$ and $N_2$ according to the ratios, $\beta$, $\beta$ and $1-2*\beta$, respectively, where $\beta\in [0, 0.5)$ is a hyperparameter. We set $\mathbf{a}_{0}=0$ for nodes in $N_0$ and $\mathbf{a}_{1}=0$ for nodes in $N_1$. Then values of $\mathbf{a}_{0}$ and $\mathbf{a}_{1}$ are kept for the remaining nodes in $N_2$.

\begin{algorithm}[t]  
	\caption{ The learning algorithm of \name }  
	\label{alg:Framwork}  
	\begin{algorithmic}[1]  
		\Require
		Adjacency matrix $\mathbf{A}$;
		Feature matrix $\mathbf{X}$;
		Training Set $V_{l}$;
		Observed labels $\mathbf{Y}_{V_{l}}$;
		\Ensure  
		Model Parameter $\Theta$ ;

		\State Generate $\mathbf{T}$ with $\mathrm{GNA}$($\mathbf{X}$, $\mathbf{A}$); \Comment{Pre-processing step}    
   
		\For{$t=1$ to $Epoch_{max}$}
		\For{$V_b$ in $V_{l}$} \Comment{Mini-batch training} 
        
        \State Calculate $\mathbf{H}^{N}_{V_{b}}$ with \autoref{eq:cnn};
        \State Calculate $\mathbf{H}^{R}_{V_{b}}$ with \autoref{eq:ra};
        
        \State Calculate the aggregation weights $\mathbf{a}_{V_{b}}$ with \autoref{eq:lwl};
        
        \State Sample $N_0$ and $N_1$ from $V_b$;
        \State Set $\mathbf{a}_{N_0, 0}=0$ and $\mathbf{a}_{N_1, 1}=0$;
        
        \State Calculate $\mathbf{H}^{F}_{V_{b}}$ with \autoref{eq:fr};
        
		\State Calculate loss $\boldsymbol{L}$ with \autoref{eq:loss}
		\State Optimize $\Theta$ by minimizing $\boldsymbol{L}$ with gradient descent
		\EndFor 
		\EndFor 
		\label{code:fram:extract}  
		
		\label{code:fram:select} \\  
		\Return Model Parameter $\Theta$;  
	\end{algorithmic}  
\end{algorithm}   

\subsection{Inference and Optimization}
Following the common setting~\cite{appnp,gprgnn,bmgcn}, we apply MLP to predict the labels of nodes according to $\mathbf{H}^{F}$, such that:
\begin{equation}
	\hat{\mathbf{Y}} = \mathrm{MLP}(\mathbf{H}^{F}), 
	\label{eq:pre}
\end{equation}
where $\hat{\mathbf{Y}}\in \mathbb{R}^{n\times c}$ denotes the predicted labels. Then, we utilize the Cross-Entropy loss, which is widely used in the node classification task~\citep{gcn,appnp,gprgnn}, to calculate the loss, given that
\begin{equation}
	\boldsymbol{L} = -\sum_{v \in V_l}\mathbf{Y}_{v}\mathrm{ln}\hat{\mathbf{Y}}_{v}. 
	\label{eq:loss}
\end{equation}
The overall learning algorithm of \name is presented in Algorithm \ref{alg:Framwork}.

\subsection{Complexity Analysis}

\subsubsection{Time complexity.} 
The time complexity of \name depends on two parts, i.e., convolutional layers and fully connected layers. Their time complexities are $O(nd^{\prime}K^2)$ and $O(ndd^{\prime}+ nd^{\prime}c)$, respectively, where $n$ is the number of nodes, $K$ the propagation step, $c$ the number of labels, $d$ and $d^{\prime}$ denote the dimension of the raw feature vector and the hidden feature vector, respectively. Thus, the total complexity of \name is $O(nd^{\prime}K^2 + ndd^{\prime}+ nd^{\prime}c)$. Since the values of $K$ and $c$ are always very small, the time complexity can be written as $O(ndd^{\prime})$. Comparing to the decoupled GCN, which has $O(mKc + ndc)$ given that $m$ is the number of edges, \name is more efficient to perform on dense graphs in which the number of edges is significantly larger than the number of nodes. 
The reason is that the time complexity of decoupled GCN relies on both $n$ and $m$, while the time complexity of \name depends only on $n$.

\subsubsection{Space complexity.}
The space complexity of \name depends on two parts, i.e., model parameters and intermediate variables. 
Their space complexities are $O(K+dd^{\prime}+d^{\prime}c)$ and $O(bKd^{\prime})$, respectively,
and the total space complexity of \name is $O(bKd^{\prime}+dd^{\prime}+d^{\prime}c)$, where $b$ denotes the batch size.
Compared to the decoupled GCN, which has $O(nc + dc + m)$, \name is more scalable on large-scale graphs as the space cost of \name does not count the graph size, $n$ and $m$.

\section{Experiments}
\begin{table}[t]
    \centering
	\caption{The statistics of datasets.}
	\label{tab:dataset}
	\scalebox{0.9}{
	\begin{tabular}{lrrrrrr}
		\toprule
		Dataset & \# Nodes & \# Edges & \# Features & \# Classes & $\mathcal{H}$\\
		\midrule
		Cora & 2,708 & 5,278 & 1,433 & 7 & 0.83\\
		Citeseer & 3,327 & 4,552 & 3,703 & 6 & 0.72\\
		Pubmed & 19,717 & 44,324 & 500 & 3 & 0.79\\
		Computer & 13,752 & 491,722 & 767 & 10 & 0.80\\
		Photo & 7,650 & 238,163 & 745 & 8 & 0.85\\

		Actor & 7,600 & 26,659 & 932 & 5 & 0.22\\
		Squirrel & 5,201 & 198,353 & 2,089 & 5 & 0.22\\
		Chameleon & 2,277 & 31,371 & 2,325 & 5 & 0.23\\
		Cornell & 183 & 277 & 1,703 & 5 & 0.30\\
		Texas & 183 & 279 & 1,703 & 5 & 0.11\\
		\bottomrule
	\end{tabular}
	}
\end{table}

\begin{table*}[t]
\small
    \centering
	\caption{
	Comparison of all models in terms of mean accuracy $\pm$ stdev (\%) on all datasets. 
	The best results appear in bold and the runner-up results are underlined.
	}
	\label{tab:all-graph}
	\setlength\tabcolsep{1mm}{
 	\scalebox{0.98}{
	\begin{tabular}{lcccccccccc}
		\toprule
		  &Cora &Citeseer &Pubmed &Computer &Photo&Actor &Squirrel &Chameleon &Cornell &Texas \\
		$\mathcal{H}(G)$&0.83 &0.72 &0.79 &0.80 &0.85&0.22 &0.22 &0.23 &0.30 &0.11 \\
		\midrule
		MLP & 77.30 $\pm$ 0.29  & 75.42 $\pm$ 0.32 & 86.92 $\pm$ 0.16  & 77.48 $\pm$ 0.28 & 86.69 $\pm$ 0.31&
		36.54 $\pm$ 0.89  & 29.77 $\pm$ 1.72 &  48.21 $\pm$ 2.98 & 82.18 $\pm$ 3.13 & 81.81 $\pm$ 3.75\\
		GCN &87.22 $\pm$ 0.41  & 77.05 $\pm$ 0.34 & 88.21 $\pm$ 0.11  & 89.53 $\pm$ 0.41 & 94.76 $\pm$ 0.17&31.52 $\pm$ 0.35  & 44.76 $\pm$ 1.21 &  59.45 $\pm$ 1.92 & 60.28 $\pm$ 2.23 & 66.28 $\pm$ 1.55\\
		GAT &88.77 $\pm$ 0.27  & 77.10 $\pm$ 0.38 & 87.87 $\pm$ 0.13  & 90.76 $\pm$ 0.56 & 94.98 $\pm$ 0.24& 29.91 $\pm$ 0.26  & 36.23 $\pm$ 1.96 &  49.71 $\pm$ 1.86 & 52.53 $\pm$ 2.08 & 62.12 $\pm$ 1.79\\
		SGC &87.59 $\pm$ 0.37  & 76.95 $\pm$ 0.19 & 87.42 $\pm$ 0.16  & 90.22 $\pm$ 0.21 & 95.11 $\pm$ 0.09&29.21 $\pm$ 0.22  & 39.63 $\pm$ 2.06 &  46.94 $\pm$ 1.85 & 62.85 $\pm$ 2.98 & 65.71 $\pm$ 3.15\\
		APPNP &89.12 $\pm$ 0.28  & 77.11 $\pm$ 0.39 & 88.69 $\pm$ 0.25  & \underline{90.91 $\pm$ 0.32} & 95.09 $\pm$ 0.14 & 34.86 $\pm$ 0.45  & 32.55 $\pm$ 1.32 &  47.25 $\pm$ 2.11 & 68.57 $\pm$ 2.02 & 71.28 $\pm$ 1.95\\

		GPR-GNN &88.96 $\pm$ 0.23  & \underline{77.30 $\pm$ 0.29} & 89.64 $\pm$ 0.34  & 89.32 $\pm$ 0.29 & \underline{95.49 $\pm$ 0.14} &36.31 $\pm$ 0.25  & 34.75 $\pm$ 1.38 &  68.54 $\pm$ 1.72 & 89.71 $\pm$ 1.01 & \underline{90.85 $\pm$ 1.50}\\
		LINKX&85.42 $\pm$ 0.43  & 74.08 $\pm$ 0.52 & 86.14 $\pm$ 0.17  & 87.35 $\pm$ 0.42 & 93.24 $\pm$ 0.24& 36.79 $\pm$ 0.85  & \underline{61.68 $\pm$ 2.97 }&  65.53 $\pm$ 2.82 & 74.58 $\pm$ 3.13 & 71.32 $\pm$ 2.95\\
		BM-GCN &\underline{89.49 $\pm$ 0.15}  & \textbf{77.47 $\pm$ 0.37} & \underline{89.82 $\pm$ 0.52 } & 89.19 $\pm$ 0.25 & 95.35 $\pm$ 0.07&34.53 $\pm$ 0.12  & 47.53 $\pm$ 1.35 &  69.12 $\pm$ 1.49 & 84.57 $\pm$ 0.83 & 89.13 $\pm$ 0.88\\
		GloGNN &88.57 $\pm$ 0.42  & 77.23 $\pm$ 0.41 & 88.58 $\pm$ 0.41  & 89.12 $\pm$ 0.33 & 95.17 $\pm$ 0.18 & \textbf{39.59 $\pm$ 0.29}  & 59.75 $\pm$ 1.81 &  \underline{75.16 $\pm$ 1.42} & \underline{91.42 $\pm$ 1.21} & 90.56 $\pm$ 1.15\\
		\name &\textbf{89.51 $\pm$ 0.14} & 77.26 $\pm$ 0.16 & \textbf{90.94 $\pm$ 0.36}  & \textbf{91.23 $\pm$ 0.13} & \textbf{95.93 $\pm$ 0.08} & \underline{38.45 $\pm$ 0.48 } & \textbf{74.33 $\pm$ 1.50} &  \textbf{78.61 $\pm$ 1.20} & \textbf{91.71 $\pm$ 1.94} & \textbf{91.14 $\pm$ 1.23}\\
		\bottomrule
	\end{tabular}}
 	}
\end{table*}

In this section, we conduct extensive experiments to evaluate the performance of \name compared with the state-of-the-art methods for node classification on ten widely used datasets. We also provide the parameter sensitivity analysis and ablation study for a deeper understanding of the proposed method.

\subsection{Experimental Setup}
\textbf{Datasets.} 
We utilize ten widely used datasets, including homophilic graphs and heterophilic graphs for experiments. The homophilic graphs contain Cora, Citeseer, Pubmed, Computers, and Photo~\citep{gprgnn}, and the heterophilic graphs include Actor, Squirrel, Chameleon, Texas, and Cornell ~\citep{geomgcn}. The statistics of datasets are reported in Table \ref{tab:dataset}, where the edge homophily ratio $\mathcal{H}$ measures the degree of homophily as introduced in Section \ref{hom}.

For the homophilic graphs, Cora, Citeseer and Pubmed are constructed from the citation networks, where nodes represent papers, and edges represent citation relationships between papers. 
Computer and Photo are extracted from the Amazon co-purchase graph, where nodes represent goods and edges indicate that two goods are frequently bought together.

For the heterophilic graphs, Actor, Squirrel and Chameleon derive from the Wikipedia pages. 
Actor is the actor-only induced subgraph of the film-director-actor-writer network where nodes correspond to actors, and the edge between two nodes denotes co-occurrence on the same Wikipedia.
Squirrel and Chameleon are page-page networks on specific topics in Wikipedia where nodes represent web pages, and edges are mutual links between pages.
Cornell and Texas are web page datasets collected from computer science departments of various universities, where nodes represent web pages and edges are hyperlinks between them.

\textbf{Baselines.}
We utilize eight state-of-the-art methods of node classification as the baselines for comparison, which are GCN~\citep{gcn}, GAT~\citep{gat}, SGC~\citep{sgcn}, APPNP~\citep{appnp}, LINKX~\citep{linkx}, GPR-GNN~\citep{gprgnn}, BM-GCN~\citep{bmgcn}, and GloGNN~\citep{glognn}. Note that the first four methods apply to homophilic graphs, while the latter four methods are designed for homophilic graphs and heterophilic graphs simultaneously. 
In addition, we also use MLP as a simple comparison method that only utilizes the attribute features of nodes to predict the node labels.

\textbf{Settings.}
We randomly split each dataset into 60\%/20\%/20\% train/val/test three parts for experiments.
For the baselines, we use their default hyper-parameters as presented in their papers, and 
for \name, we use the grid search method to determine the hyper-parameters. 
We try the hidden dimension in $\{128, 256, 512\}$, 
$\beta$ in $\{0.1, 0.2, 0.3, 0.4\}$,
and the propagation steps in $\{2,4,\cdots,10\}$.
Parameters are optimized with AdamW~\citep{adamw} optimizer, with the learning rate of in $\{1e-3,5e-4,1e-4,\}$ and weight decay in $\{1e-4,1e-5\}$. 
The batch size is set to 1000.
The training process is early stopped within 50 epochs.
All experiments are conducted on a Linux server with 1 I9-9900k CPU, 1 RTX 2080TI GPU and 64G RAM. 

\begin{figure*}[]
    \centering
	\includegraphics[width=6in]{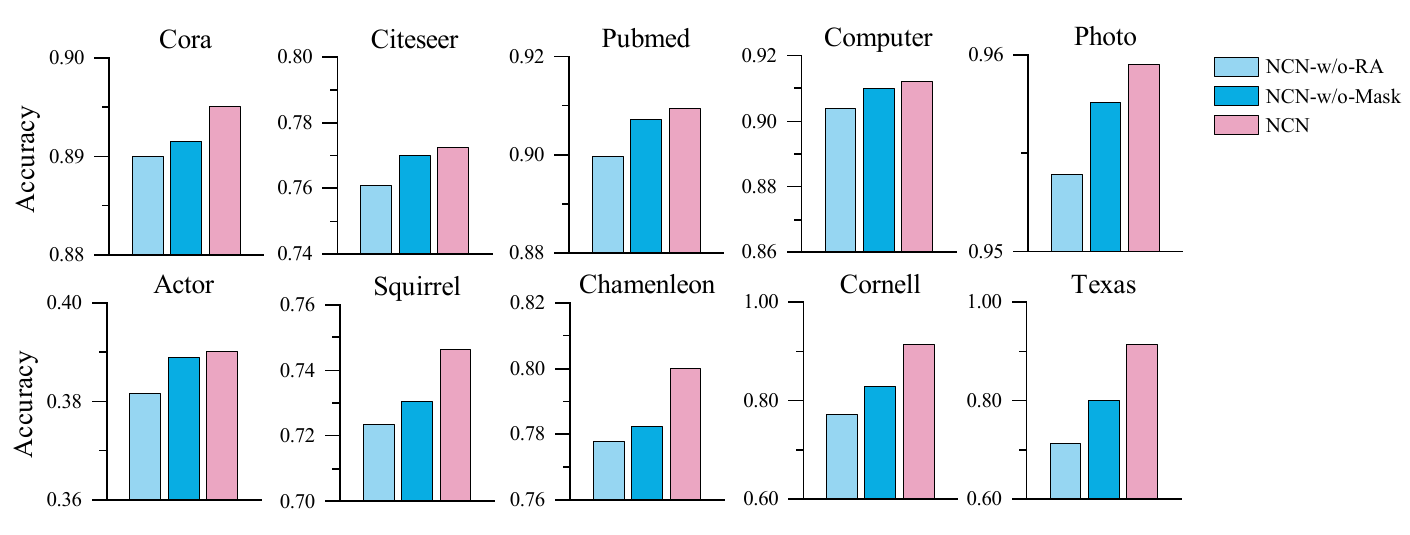}
	\caption{The performances of \name and its variants on all datasets. }
	\label{fig:as}
\end{figure*}

\subsection{Performance Comparison}
We adopt accuracy as the evaluation metric for the node classification task to compare the model performance.
We run ten times for each method with random splits and report the average results as summarized in \autoref{tab:all-graph}.
We can observe that our \name achieves the best performance on almost all the datasets except Citeseer and Actor.
Particularly, \name outperforms GloGNN, BM-GCN and LINKX, which are powerful methods for node classification on heterophilic graphs, demonstrating the effectiveness of \name.
Although \name misses the best results on Citeseer and Actor, it still exhibits competitive performances on these datasets.
The results confirm that \name is capable of the node classification task on both homophilic and heterophilic graphs, showing the effectiveness of \name.
Moreover, \name beats representative decoupled GCNs (APPNP, SGC and GPRGNN) on all the datasets (except behind GPRGNN on Citeseer), indicating that the designs of \name are beneficial for learning the powerful node representations from the information of multi-hop neighborhoods.

\subsection{Ablation Study}
We compare \name with its two variants on all datasets for evaluating the effectiveness of the key designs in \name. 
The two variants of \name are described as follows
\begin{itemize}
    \item \textbf{\name-w/o-RA}: We remove the adaptive feature fusion from \name, and only the features learned from the neighborhood information is preserved.
    \item \textbf{\name-w/o-Mask}: We do not utilize the mask training strategy in this variant.
\end{itemize}

According to the comparison results in \autoref{fig:as}, we can find that \name outperforms the variants on all datasets, demonstrating that the proposed feature fusion module and mask-based training strategy can improve the performance for the node classification task. Furthermore, \name-w/o-Mask beats \name-w/o-RA on all datasets, indicating that introducing the original features of nodes plays a vital role in learning meaningful representations of the nodes.

\subsection{Parameter Analysis}
\begin{figure*}[t]
    \centering
	\includegraphics[width=7in]{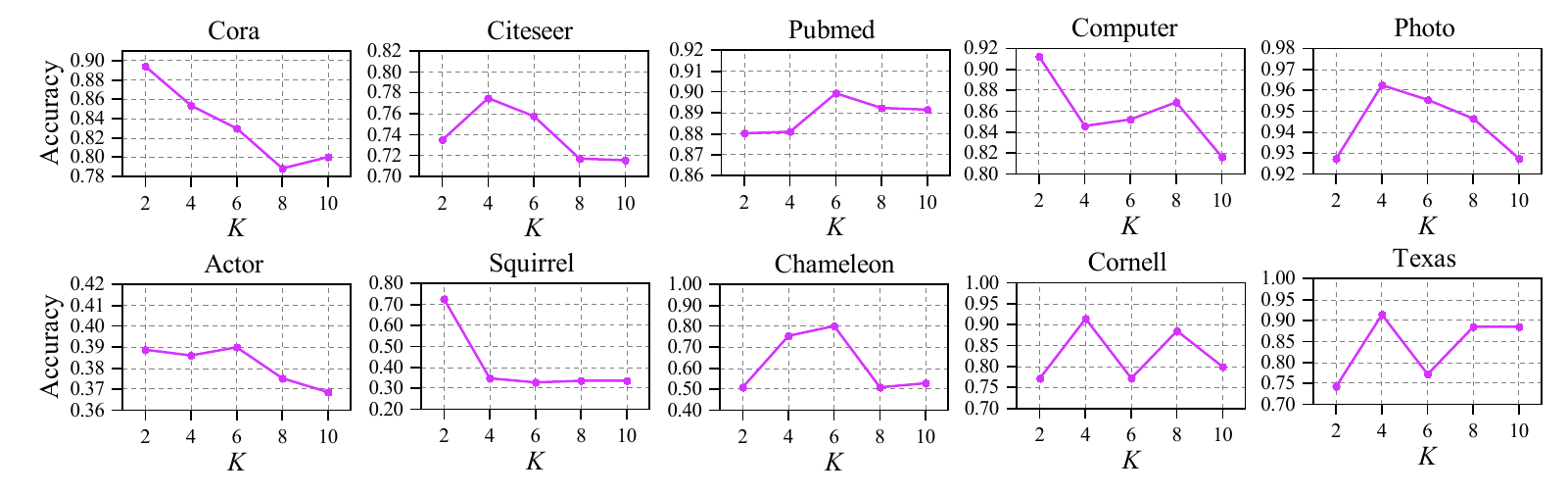}
	\caption{ The influence of the propagation step $K$ on the model performance.}
	\label{fig:k}
\end{figure*}

In this subsection, we perform experiments on four datasets to explore the influence of the propagation step $K$ on the performance of \name.
We vary the value of $K$ in $\{2,4,6,8,10\}$ and give the corresponding results in \autoref{fig:k}.
We observe that $K$ affects \name to a large extent because it determines the range of aggregated information for nodes. For different datasets, the value of $K$ varies for the best performance of \name because the characteristics of graphs are different. Furthermore, we can also find that setting $K$ less than or equal to 6 could obtain good results. It is because the aggregated information involves too much irrelevant information when $K$ is large that weakens the performance of \name.

\subsection{Visualization}
\label{sec:visual}
In this subsection, to intuitively demonstrate the effectiveness of our proposed adaptive feature fusion model, we visualize the distribution of the learnable aggregation weights, $a_0$ and $a_1$ in \autoref{eq:fr}, which measure the importance of the information learned from the node itself and its neighborhoods respectively.
We select four representative datasets, including Pubmed, Photo, Squirrel and Chameleon. According to the results of MLP in \autoref{tab:all-graph}, the raw attribute features of nodes are highly correlated to the labels of nodes on Pubmed and Photo. In contrast, the raw attribute features of nodes are not quite irrelevant to the labels of nodes on Squirrel and Chameleon.
We randomly select 100 nodes from the test set for each dataset and show the distributions of aggregation weights in \autoref{fig:as}.

In \autoref{fig:as}, we can find that the average values of $a_0$ and $a_1$ are similar on Pubmed and Photo indicating that the raw attribute features and multi-hop neighborhoods are almost equally important for learning the node representations. On the other hand, the values of $a_1$ are extremely larger
than that of $a_0$ on Squirrel and Chameleon, demonstrating that the information of neighborhoods plays a more important role in learning the representations of nodes than the raw attribute features. Therefore, we can conclude that 
our proposed method can adaptively and effectively learn the aggregation weights according to the characteristics of different graphs.

\begin{figure}[t]
    \centering
	\includegraphics[width=3.2in]{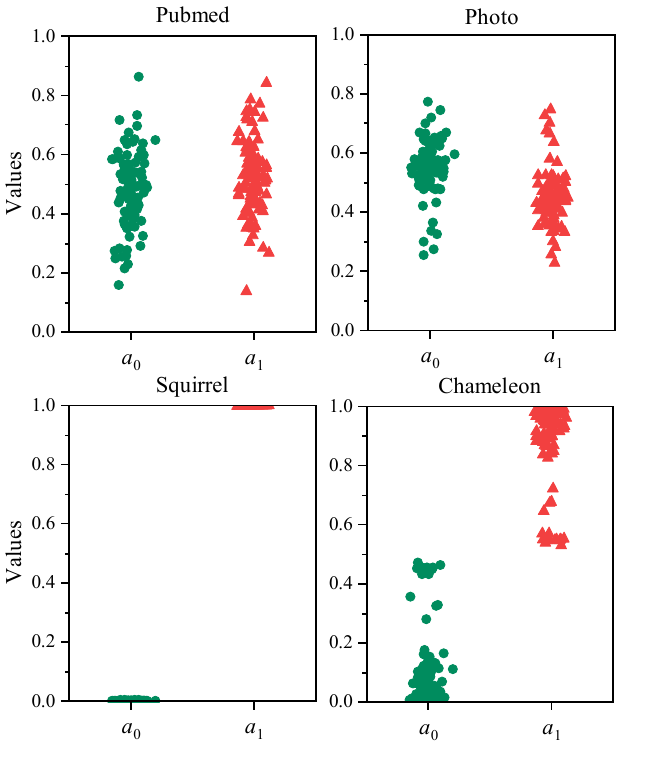}
	\caption{The distributions of the aggregation weights on different datasets.}
	\label{fig:as}
\end{figure}

\section{Related Work}
This section reviews recent works of GCN and decoupled GCN.

\textbf{Graph Convolutional Network.} 
Graph Convolutional Network (GCN)~\cite{gcn} is a typical GNN that applies a non-linear smoothing operation on the first-order neighbors of nodes to learn the representations of nodes.
Due to its flexibility, many follow-up GCN models have been proposed to improve performance.
In the original GCN, the aggregation operation only relies on the graph topology. The attribute information, one of the most important information of graphs, is ignored, preventing the model from learning the powerful node representations.
To address this problem, some recent works~\cite{gat,amgcn,sim-gcn,ugcn} introduce the attribute features into the aggregation of the plain GCN for preserving both topology and attribute information.
GAT~\cite{gat} introduces the attention mechanism to specify different weights to different one-hop neighbors of the central node for aggregating the information adaptively.
AM-GCN~\cite{amgcn} utilizes the GCN layer to extract the features from the original graph and similarity graph generated by node attributes to learn the node representations from the perspectives of topology information and attribute information.

Recent studies argue that GCN implicitly obeys the graph homophily assumption~\citep{h2gnn, glognn} that the connected nodes tend to share the same labels. Thus, GCN performs poorly on heterophilic graphs where adjacent nodes tend to have different labels.
Many works have been proposed to overcome this limitation in recent years~\cite{heter-gnns-survey}.
Zhu~\etal~\cite{h2gnn} theoretically prove that two-hop neighbors tend to contain more nodes with the same class of the central node and propose H2GNN, which combines the features of one-hop neighborhood and two-hop neighborhood in a GCN layer to strengthen the representation of the central node.
BM-GCN~\cite{bmgcn} and HOG-GCN~\cite{hoggcn} generate the soft labels of unlabeled nodes based on the raw attribute information. And they further estimate the block matrix or homophily degree matrix to learn the class-aware similarity of node pairs for guiding the aggregation operation of GCN. 
For more works of generalizing GCN to heterophilic graphs, please refer to a survey~\cite{heter-gnns-survey}.

\textbf{Decoupled Graph Convolutional Network.} 
The coupled design of neighborhood aggregation and feature transformation in the GCN layer will increase the cost of model training~\citep{appnp, gbp} and cause the over-smoothing issue~\cite{deepgcn0, deepgcn1, deepergcn} when stacking many GCN layers, limiting the model performance. 
Recent works, called decoupled GCNs~\cite{sgcn,appnp,gprgnn,deepergcn,gbp}, decouple the neighborhood aggregation and feature transformation and treat them as independent modules, and have become the latest paradigm of GCN~\cite{pt}.
Some works introduce various propagation methods for aggregating neighborhood information, such as random walk~\cite{sgcn} and personalized PageRank~\cite{appnp}.
To reduce the computational cost of the propagation operation, Chen~\etal~\cite{gbp} developed a localized bidirectional propagation process to accelerate the model training.
Several works aim to design various feature extractors for learning the node presentations from multi-hop neighborhoods.
GPRGNN~\cite{gprgnn} assigns learnable aggregation weights to each hop neighborhood to fuse the neighborhood information adaptively. 
DAGNN~\cite{deepergcn} utilizes a trainable projection vector to adaptively balance the information from local and global neighborhoods for each node.

Despite effectiveness, 
the feature extractors in decoupled GCN could be regarded as a simple linear combination, limiting the model's ability to capture more meaningful information from the multi-hop neighborhoods. 
In contrast, our proposed \name abstracts the feature extractor as an independent and flexible module, enabling the model to develop more powerful neural network-based feature extractors for capturing the complex semantic features of multi-hop neighborhoods.

\section{Conclusion}

In this paper, we propose a general framework, termed Neighborhood Convolutional Network (\name), which is a new paradigm of Graph Convolutional Neural Network for the node classification task.
Benefiting from the proposed pre-processing aggregator that transforms the neighborhood features into the grid-like feature matrix, \name develops a powerful CNN based module to extract the neighborhood features from the generated feature matrix of multi-hop neighborhoods.
Through an adaptive feature fusion module, \name can effectively learn the node representations from the raw attribute features and neighborhood features adaptively.
Moreover, a mask-based training strategy is developed to further boost the model's performance.
We evaluate \name on ten benchmark datasets, including homophilic and heterophilic graphs.
The results demonstrate the superiority of \name over the state-of-the-art baselines for node classification.

In future work, we will further explore the \name model in two main directions. First, we will apply other possible model designs to extract the neighborhood information, such as ResNet~\cite{resnet}. Second, we will investigate the framework of \name on other well-known tasks, such as link prediction and graph classification.

\bibliographystyle{ACM-Reference-Format}
\bibliography{sample-base}


\begin{thebibliography}{28}


\ifx \showCODEN    \undefined \def \showCODEN     #1{\unskip}     \fi
\ifx \showDOI      \undefined \def \showDOI       #1{#1}\fi
\ifx \showISBNx    \undefined \def \showISBNx     #1{\unskip}     \fi
\ifx \showISBNxiii \undefined \def \showISBNxiii  #1{\unskip}     \fi
\ifx \showISSN     \undefined \def \showISSN      #1{\unskip}     \fi
\ifx \showLCCN     \undefined \def \showLCCN      #1{\unskip}     \fi
\ifx \shownote     \undefined \def \shownote      #1{#1}          \fi
\ifx \showarticletitle \undefined \def \showarticletitle #1{#1}   \fi
\ifx \showURL      \undefined \def \showURL       {\relax}        \fi
\providecommand\bibfield[2]{#2}
\providecommand\bibinfo[2]{#2}
\providecommand\natexlab[1]{#1}
\providecommand\showeprint[2][]{arXiv:#2}

\bibitem[Chen et~al\mbox{.}(2020a)]%
        {deepgcn1}
\bibfield{author}{\bibinfo{person}{Deli Chen}, \bibinfo{person}{Yankai Lin},
  \bibinfo{person}{Wei Li}, \bibinfo{person}{Peng Li}, \bibinfo{person}{Jie
  Zhou}, {and} \bibinfo{person}{Xu Sun}.} \bibinfo{year}{2020}\natexlab{a}.
\newblock \showarticletitle{Measuring and Relieving the Over-Smoothing Problem
  for Graph Neural Networks from the Topological View}. In
  \bibinfo{booktitle}{\emph{Proceedings of the {AAAI} Conference on Artificial
  Intelligence, 2020}}.
\newblock


\bibitem[Chen et~al\mbox{.}(2020b)]%
        {gbp}
\bibfield{author}{\bibinfo{person}{Ming Chen}, \bibinfo{person}{Zhewei Wei},
  \bibinfo{person}{Bolin Ding}, \bibinfo{person}{Yaliang Li},
  \bibinfo{person}{Ye Yuan}, \bibinfo{person}{Xiaoyong Du}, {and}
  \bibinfo{person}{Ji-Rong Wen}.} \bibinfo{year}{2020}\natexlab{b}.
\newblock \showarticletitle{Scalable Graph Neural Networks via Bidirectional
  Propagation}. In \bibinfo{booktitle}{\emph{Proceedings of the Advances in
  Neural Information Processing Systems, 2020}}.
\newblock


\bibitem[Chen et~al\mbox{.}(2020c)]%
        {gcnii}
\bibfield{author}{\bibinfo{person}{Ming Chen}, \bibinfo{person}{Zhewei Wei},
  \bibinfo{person}{Zengfeng Huang}, \bibinfo{person}{Bolin Ding}, {and}
  \bibinfo{person}{Yaliang Li}.} \bibinfo{year}{2020}\natexlab{c}.
\newblock \showarticletitle{Simple and Deep Graph Convolutional Networks}. In
  \bibinfo{booktitle}{\emph{Proceedings of the International Conference on
  Machine Learning, 2020}}.
\newblock


\bibitem[Chien et~al\mbox{.}(2021)]%
        {gprgnn}
\bibfield{author}{\bibinfo{person}{Eli Chien}, \bibinfo{person}{Jianhao Peng},
  \bibinfo{person}{Pan Li}, {and} \bibinfo{person}{Olgica Milenkovic}.}
  \bibinfo{year}{2021}\natexlab{}.
\newblock \showarticletitle{Adaptive Universal Generalized PageRank Graph
  Neural Network}. In \bibinfo{booktitle}{\emph{Proceedings of the
  International Conference on Learning Representations, 2021}}.
\newblock


\bibitem[Defferrard et~al\mbox{.}(2016)]%
        {sp-gcn}
\bibfield{author}{\bibinfo{person}{Micha{\"{e}}l Defferrard},
  \bibinfo{person}{Xavier Bresson}, {and} \bibinfo{person}{Pierre
  Vandergheynst}.} \bibinfo{year}{2016}\natexlab{}.
\newblock \showarticletitle{Convolutional Neural Networks on Graphs with Fast
  Localized Spectral Filtering}. In \bibinfo{booktitle}{\emph{Proceedings of
  the Advances in Neural Information Processing Systems, 2016}}.
\newblock


\bibitem[Dong et~al\mbox{.}(2021)]%
        {pt}
\bibfield{author}{\bibinfo{person}{Hande Dong}, \bibinfo{person}{Jiawei Chen},
  \bibinfo{person}{Fuli Feng}, \bibinfo{person}{Xiangnan He},
  \bibinfo{person}{Shuxian Bi}, \bibinfo{person}{Zhaolin Ding}, {and}
  \bibinfo{person}{Peng Cui}.} \bibinfo{year}{2021}\natexlab{}.
\newblock \showarticletitle{On the Equivalence of Decoupled Graph Convolution
  Network and Label Propagation}. In \bibinfo{booktitle}{\emph{Proceedings of
  the Web Conference, 2021}}.
\newblock


\bibitem[Gilmer et~al\mbox{.}(2017)]%
        {mpnn}
\bibfield{author}{\bibinfo{person}{Justin Gilmer}, \bibinfo{person}{Samuel~S
  Schoenholz}, \bibinfo{person}{Patrick~F Riley}, \bibinfo{person}{Oriol
  Vinyals}, {and} \bibinfo{person}{George~E Dahl}.}
  \bibinfo{year}{2017}\natexlab{}.
\newblock \showarticletitle{Neural Message Passing for Quantum Chemistry}. In
  \bibinfo{booktitle}{\emph{Proceedings of the International Conference on
  Machine Learning, 2017}}.
\newblock


\bibitem[He et~al\mbox{.}(2022)]%
        {bmgcn}
\bibfield{author}{\bibinfo{person}{Dongxiao He}, \bibinfo{person}{Chundong
  Liang}, \bibinfo{person}{Huixin Liu}, \bibinfo{person}{Mingxiang Wen},
  \bibinfo{person}{Pengfei Jiao}, {and} \bibinfo{person}{Zhiyong Feng}.}
  \bibinfo{year}{2022}\natexlab{}.
\newblock \showarticletitle{Block Modeling-Guided Graph Convolutional Neural
  Networks}. In \bibinfo{booktitle}{\emph{Proceedings of the {AAAI} Conference
  on Artificial Intelligence, 2022}}.
\newblock


\bibitem[He et~al\mbox{.}(2016)]%
        {resnet}
\bibfield{author}{\bibinfo{person}{Kaiming He}, \bibinfo{person}{Xiangyu
  Zhang}, \bibinfo{person}{Shaoqing Ren}, {and} \bibinfo{person}{Jian Sun}.}
  \bibinfo{year}{2016}\natexlab{}.
\newblock \showarticletitle{Deep Residual Learning for Image Recognition}. In
  \bibinfo{booktitle}{\emph{Proceedings of the {IEEE} Conference on Computer
  Vision and Pattern Recognition, 2016}}.
\newblock


\bibitem[Jin et~al\mbox{.}(2021b)]%
        {ugcn}
\bibfield{author}{\bibinfo{person}{Di Jin}, \bibinfo{person}{Zhizhi Yu},
  \bibinfo{person}{Cuiying Huo}, \bibinfo{person}{Rui Wang},
  \bibinfo{person}{Xiao Wang}, \bibinfo{person}{Dongxiao He}, {and}
  \bibinfo{person}{Jiawei Han}.} \bibinfo{year}{2021}\natexlab{b}.
\newblock \showarticletitle{Universal Graph Convolutional Networks}. In
  \bibinfo{booktitle}{\emph{Proceedings of the Advances in Neural Information
  Processing Systems, 2021}}.
\newblock


\bibitem[Jin et~al\mbox{.}(2021a)]%
        {sim-gcn}
\bibfield{author}{\bibinfo{person}{Wei Jin}, \bibinfo{person}{Tyler Derr},
  \bibinfo{person}{Yiqi Wang}, \bibinfo{person}{Yao Ma}, \bibinfo{person}{Zitao
  Liu}, {and} \bibinfo{person}{Jiliang Tang}.}
  \bibinfo{year}{2021}\natexlab{a}.
\newblock \showarticletitle{Node similarity preserving graph convolutional
  networks}. In \bibinfo{booktitle}{\emph{Proceedings of the ACM International
  Conference on Web Search and Data Mining, 2021}}.
\newblock


\bibitem[Kipf and Welling(2017)]%
        {gcn}
\bibfield{author}{\bibinfo{person}{Thomas~N. Kipf} {and} \bibinfo{person}{Max
  Welling}.} \bibinfo{year}{2017}\natexlab{}.
\newblock \showarticletitle{Semi-Supervised Classification with Graph
  Convolutional Networks}. In \bibinfo{booktitle}{\emph{Proceedings of the
  International Conference on Learning Representations, 2017}}.
\newblock


\bibitem[Klicpera et~al\mbox{.}(2019a)]%
        {appnp}
\bibfield{author}{\bibinfo{person}{Johannes Klicpera},
  \bibinfo{person}{Aleksandar Bojchevski}, {and} \bibinfo{person}{Stephan
  G{\"{u}}nnemann}.} \bibinfo{year}{2019}\natexlab{a}.
\newblock \showarticletitle{Predict then Propagate: Graph Neural Networks meet
  Personalized PageRank}. In \bibinfo{booktitle}{\emph{Proceedings of the
  International Conference on Learning Representations, 2019}}.
\newblock


\bibitem[Klicpera et~al\mbox{.}(2019b)]%
        {gdc}
\bibfield{author}{\bibinfo{person}{Johannes Klicpera}, \bibinfo{person}{Stefan
  Wei{\ss}enberger}, {and} \bibinfo{person}{Stephan G{\"u}nnemann}.}
  \bibinfo{year}{2019}\natexlab{b}.
\newblock \showarticletitle{Diffusion improves graph learning}. In
  \bibinfo{booktitle}{\emph{Proceedings of the Advances in Neural Information
  Processing Systems, 2019}}.
\newblock


\bibitem[Li et~al\mbox{.}(2018)]%
        {deepgcn0}
\bibfield{author}{\bibinfo{person}{Qimai Li}, \bibinfo{person}{Zhichao Han},
  {and} \bibinfo{person}{Xiao{-}Ming Wu}.} \bibinfo{year}{2018}\natexlab{}.
\newblock \showarticletitle{Deeper Insights Into Graph Convolutional Networks
  for Semi-Supervised Learning}. In \bibinfo{booktitle}{\emph{Proceedings of
  the {AAAI} Conference on Artificial Intelligence, 2018}}.
\newblock


\bibitem[Li et~al\mbox{.}(2022)]%
        {glognn}
\bibfield{author}{\bibinfo{person}{Xiang Li}, \bibinfo{person}{Renyu Zhu},
  \bibinfo{person}{Yao Cheng}, \bibinfo{person}{Caihua Shan},
  \bibinfo{person}{Siqiang Luo}, \bibinfo{person}{Dongsheng Li}, {and}
  \bibinfo{person}{Weining Qian}.} \bibinfo{year}{2022}\natexlab{}.
\newblock \showarticletitle{Finding Global Homophily in Graph Neural Networks
  When Meeting Heterophily}.
\newblock \bibinfo{journal}{\emph{arXiv preprint arXiv:2205.07308}}
  (\bibinfo{year}{2022}).
\newblock


\bibitem[Lim et~al\mbox{.}(2021)]%
        {linkx}
\bibfield{author}{\bibinfo{person}{Derek Lim}, \bibinfo{person}{Felix Hohne},
  \bibinfo{person}{Xiuyu Li}, \bibinfo{person}{Sijia~Linda Huang},
  \bibinfo{person}{Vaishnavi Gupta}, \bibinfo{person}{Omkar Bhalerao}, {and}
  \bibinfo{person}{Ser~Nam Lim}.} \bibinfo{year}{2021}\natexlab{}.
\newblock \showarticletitle{Large scale learning on non-homophilous graphs: New
  benchmarks and strong simple methods}. In
  \bibinfo{booktitle}{\emph{Proceedings of the Advances in Neural Information
  Processing Systems, 2021}}.
\newblock


\bibitem[Liu et~al\mbox{.}(2020)]%
        {deepergcn}
\bibfield{author}{\bibinfo{person}{Meng Liu}, \bibinfo{person}{Hongyang Gao},
  {and} \bibinfo{person}{Shuiwang Ji}.} \bibinfo{year}{2020}\natexlab{}.
\newblock \showarticletitle{Towards Deeper Graph Neural Networks}. In
  \bibinfo{booktitle}{\emph{Proceedings of the {ACM} {SIGKDD} Conference on
  Knowledge Discovery and Data Mining, 2020}}.
\newblock


\bibitem[Loshchilov and Hutter(2019)]%
        {adamw}
\bibfield{author}{\bibinfo{person}{Ilya Loshchilov} {and}
  \bibinfo{person}{Frank Hutter}.} \bibinfo{year}{2019}\natexlab{}.
\newblock \showarticletitle{Decoupled Weight Decay Regularization}. In
  \bibinfo{booktitle}{\emph{Proceedings of the International Conference on
  Learning Representations, 2019}}.
\newblock


\bibitem[Pei et~al\mbox{.}(2020)]%
        {geomgcn}
\bibfield{author}{\bibinfo{person}{Hongbin Pei}, \bibinfo{person}{Bingzhe Wei},
  \bibinfo{person}{Kevin~Chen{-}Chuan Chang}, \bibinfo{person}{Yu Lei}, {and}
  \bibinfo{person}{Bo Yang}.} \bibinfo{year}{2020}\natexlab{}.
\newblock \showarticletitle{Geom-GCN: Geometric Graph Convolutional Networks}.
  In \bibinfo{booktitle}{\emph{Proceedings of the International Conference on
  Learning Representations, 2020}}.
\newblock


\bibitem[Srivastava et~al\mbox{.}(2014)]%
        {dropout}
\bibfield{author}{\bibinfo{person}{Nitish Srivastava},
  \bibinfo{person}{Geoffrey Hinton}, \bibinfo{person}{Alex Krizhevsky},
  \bibinfo{person}{Ilya Sutskever}, {and} \bibinfo{person}{Ruslan
  Salakhutdinov}.} \bibinfo{year}{2014}\natexlab{}.
\newblock \showarticletitle{Dropout: a simple way to prevent neural networks
  from overfitting}.
\newblock \bibinfo{journal}{\emph{The Journal of Machine Learning Research}}
  \bibinfo{volume}{15}, \bibinfo{number}{1} (\bibinfo{year}{2014}),
  \bibinfo{pages}{1929--1958}.
\newblock


\bibitem[Velikovi et~al\mbox{.}(2018)]%
        {gat}
\bibfield{author}{\bibinfo{person}{Petar Velikovi}, \bibinfo{person}{G.
  Cucurull}, \bibinfo{person}{A. Casanova}, \bibinfo{person}{A. Romero},
  \bibinfo{person}{P Liò}, {and} \bibinfo{person}{Y. Bengio}.}
  \bibinfo{year}{2018}\natexlab{}.
\newblock \showarticletitle{Graph Attention Networks}. In
  \bibinfo{booktitle}{\emph{Proceedings of the International Conference on
  Learning Representations, 2018}}.
\newblock


\bibitem[Wang et~al\mbox{.}(2021)]%
        {hoggcn}
\bibfield{author}{\bibinfo{person}{Tao Wang}, \bibinfo{person}{Rui Wang},
  \bibinfo{person}{Di Jin}, \bibinfo{person}{Dongxiao He}, {and}
  \bibinfo{person}{Yuxiao Huang}.} \bibinfo{year}{2021}\natexlab{}.
\newblock \showarticletitle{Powerful Graph Convolutioal Networks with Adaptive
  Propagation Mechanism for Homophily and Heterophily}.
\newblock \bibinfo{journal}{\emph{CoRR}}  \bibinfo{volume}{abs/2112.13562}
  (\bibinfo{year}{2021}).
\newblock


\bibitem[Wang et~al\mbox{.}(2020)]%
        {amgcn}
\bibfield{author}{\bibinfo{person}{Xiao Wang}, \bibinfo{person}{Meiqi Zhu},
  \bibinfo{person}{Deyu Bo}, \bibinfo{person}{Peng Cui}, \bibinfo{person}{Chuan
  Shi}, {and} \bibinfo{person}{Jian Pei}.} \bibinfo{year}{2020}\natexlab{}.
\newblock \showarticletitle{{AM-GCN:} Adaptive Multi-channel Graph
  Convolutional Networks}. In \bibinfo{booktitle}{\emph{Proceedings of the
  {ACM} {SIGKDD} Conference on Knowledge Discovery and Data Mining, 2020}}.
\newblock


\bibitem[Wu et~al\mbox{.}(2019)]%
        {sgcn}
\bibfield{author}{\bibinfo{person}{Felix Wu}, \bibinfo{person}{Amauri H.~Souza
  Jr.}, \bibinfo{person}{Tianyi Zhang}, \bibinfo{person}{Christopher Fifty},
  \bibinfo{person}{Tao Yu}, {and} \bibinfo{person}{Kilian~Q. Weinberger}.}
  \bibinfo{year}{2019}\natexlab{}.
\newblock \showarticletitle{Simplifying Graph Convolutional Networks}. In
  \bibinfo{booktitle}{\emph{Proceedings of the International Conference on
  Machine Learning, 2019}}.
\newblock


\bibitem[Xu et~al\mbox{.}(2018)]%
        {jknet}
\bibfield{author}{\bibinfo{person}{Keyulu Xu}, \bibinfo{person}{Chengtao Li},
  \bibinfo{person}{Yonglong Tian}, \bibinfo{person}{Tomohiro Sonobe},
  \bibinfo{person}{Ken-ichi Kawarabayashi}, {and} \bibinfo{person}{Stefanie
  Jegelka}.} \bibinfo{year}{2018}\natexlab{}.
\newblock \showarticletitle{Representation Learning on Graphs with Jumping
  Knowledge Networks}. In \bibinfo{booktitle}{\emph{Proceedings of the
  International conference on machine learning, 2018}}.
\newblock


\bibitem[Zheng et~al\mbox{.}(2022)]%
        {heter-gnns-survey}
\bibfield{author}{\bibinfo{person}{Xin Zheng}, \bibinfo{person}{Yixin Liu},
  \bibinfo{person}{Shirui Pan}, \bibinfo{person}{Miao Zhang},
  \bibinfo{person}{Di Jin}, {and} \bibinfo{person}{Philip~S. Yu}.}
  \bibinfo{year}{2022}\natexlab{}.
\newblock \showarticletitle{Graph Neural Networks for Graphs with Heterophily:
  {A} Survey}.
\newblock \bibinfo{journal}{\emph{CoRR}}  \bibinfo{volume}{abs/2202.07082}
  (\bibinfo{year}{2022}).
\newblock


\bibitem[Zhu et~al\mbox{.}(2020)]%
        {h2gnn}
\bibfield{author}{\bibinfo{person}{Jiong Zhu}, \bibinfo{person}{Yujun Yan},
  \bibinfo{person}{Lingxiao Zhao}, \bibinfo{person}{Mark Heimann},
  \bibinfo{person}{Leman Akoglu}, {and} \bibinfo{person}{Danai Koutra}.}
  \bibinfo{year}{2020}\natexlab{}.
\newblock \showarticletitle{Beyond Homophily in Graph Neural Networks: Current
  Limitations and Effective Designs}. In \bibinfo{booktitle}{\emph{Proceedings
  of the Advances in Neural Information Processing Systems, 2020}}.
\newblock


\end{thebibliography}


\end{document}